# Computational modeling of early language learning from acoustic speech and audiovisual input without linguistic priors

Okko Räsänen

Signal Processing Research Centre, Tampere University, Finland

## Abstract

Learning to understand speech appears almost effortless for typically developing infants, yet from an information-processing perspective, acquiring a language from acoustic speech is an enormous challenge. This chapter reviews recent developments in using computational models to understand early language acquisition from speech and audiovisual input. The focus is on self-supervised and visually grounded models of perceptual learning. We show how these models are becoming increasingly powerful in learning various aspects of speech without strong linguistic priors, and how many features of early language development can be explained through a shared set of learning principles—principles broadly compatible with multiple theories of language acquisition and human cognition. We also discuss how modern learning simulations are gradually becoming more realistic, both in terms of input data and in linking model behavior to empirical findings on infant language development.

## Introduction

The way human infants learn to understand and produce their native language appears almost effortless. Instead of receiving carefully curated speech input together with a linguistic description of its structure, infants simply learn from everyday interactions with their caregivers and the surrounding world. This gradual yet robust transformation of general sensorimotor experiences into a full-blown language system has long puzzled researchers across many fields, such as linguistics, developmental science, cognitive psychology, neuroscience, and artificial intelligence.

The more one considers the potential cognitive underpinnings of the early language learning process, the more complex it appears. To comprehend a language, the learner must be able to segment linguistic units, such as phones, syllables, or words, from the continuous stream of acoustic sensory stimulation, physically realized as temporal air pressure variations. In addition, these segments must be assigned a categorical identity (e.g., a phoneme /b/, syllable /bi/, or word "*bit*"), and the compositional role of these elements must be syntactically parsed in order to understand the message (e.g., is the word "*bit*" used as a noun or as a past tense of the verb "*bite*"). Moreover, to play any functional role[1] for the listener, the received signal must somehow be linked to its meaning in the world outside language, often referred to as grounding (e.g., Quine, 1960). Hence, the learning process can be viewed as involving segmentation, categorization, parsing, and meaning acquisition problems at various levels of linguistic structure.

Importantly, many of these learning problems are dependent on each other and are therefore difficult to solve independently or sequentially. For instance, phonemes—the elementary building blocks of language—are defined in terms of acoustic contrasts that change the meanings of words in the language (e.g., /i/ vs. /æ/ in "*bit*" and "*bat*"), and these contrasts vary across languages. At the same time, linguistics defines words as entities composed of phonemes (or morphemes composed of phonemes), but also as the smallest units of language that carry standalone meaning, that is, some relation to the world outside language. This implies that one level of elementary linguistic structure is unlikely to precede the other in acquisition; rather, there are multiple interdependencies across linguistic levels and between language and the external world that must be accounted for simultaneously. Moreover, pragmatic factors also play a role, as prosodic and syntactic cues shape how listeners infer speakers' intentions in everyday situations.

To complicate early language learning further, real speech comes with substantial acoustic variability that causes the speech and all its linguistic units to sound different on every occasion. In some temporal contexts, phoneme /e/ may sound like phone [i] instead of [e], whereas the same utterance spoken by a father has different acoustic characteristics from those of the mother (e.g., Hillenbrand et al., 1995). The mother also sounds different when speaking slowly or quickly, when tired or excited, and when addressing the child, the father, or her own mother. In addition, virtually all real-world speech input is mixed with various environmental sounds and modified by reverberations that further alter the physical form and hence the auditory imprint of the signal. In general, speech does not come with universal cues for identifying different linguistic elements. Instead, most of the linguistic structure can be viewed as latent, hidden from naïve observers and camouflaged in different ways across various languages, dialects, and speakers. Overall, the hierarchical structure with dependencies across levels, together with acoustic variability, reveals that language acquisition—though it may superficially appear easy for a child to learn—is in reality an enormously complex cognitive process.

The sheer number of different "sub-problems" in language acquisition also raises a fundamental question: why would infant learners try to solve any of the individual learning problems in isolation, when they still have to master the entire processing chain from input to meaning before the speech input can become useful to them? What are the underlying cognitive processes responsible for driving the incremental yet incredibly robust learning process—a process that uses everyday language experiences as ingredients to build increasingly functional language capabilities? Are statistical and distributional learning of input regularities sufficient (e.g., Maye et al., 2002; Saffran & Kirkham, 2018)? What kinds of innate linguistic priors or other inductive biases are required to guide the process (see, e.g., Culbertson et al., 2013)? What roles do interaction, quality and quantity of language input, or multimodality play in the process? What about speech production: is it necessary or helpful for learning to comprehend language (cf. Liberman & Mattingly, 1985)? In general, how many different constraints and learning mechanisms are needed, and how likely are they all to exist in our genetic endowment?

[1] Functional role of language refers here to the ability of speech input to somehow impact behavior of the listener, either at immediate or delayed time-scales.

Notably, the complexity of language also makes it difficult to study language learning in infants. The standard scientific approach—isolating an individual aspect of language while assuming the rest of the system as constant—struggles to capture the dependencies that exist across different aspects and levels of language and communicative behavior in humans. While many of the questions on early language development can be (and have been) tackled with observational and in-lab empirical studies, integration of individual empirical findings into mechanistic holistic theories of early language development has turned out to be challenging. Moreover, since language skills develop gradually over time and substantial individual variability exists in the developmental trajectories of infants, it is both difficult and expensive to conduct controlled studies that could precisely test integrated theories of the learning process.

As a result, several high-level theoretical frameworks have attempted to capture the main aspects of the development process in verbal terms (e.g., Jusczyk, 1993; Kuhl et al., 2007; Rowland et al., 2026; Werker & Curtin, 2005), but there are no commonly accepted "grand theories" of early language development that would be mechanistic enough to permit unambiguous empirical validation or falsification. This is where computational modeling of early language development can potentially become useful.

**Computational modeling to the rescue**

In contrast to behavioral experiments on infants, where the standard approach aims to isolate one aspect of language while controlling for others, computational models of learning *can,* and *must,* explicitly specify the entire processing chain from low-level sensory inputs to some kind of language learning outcomes (see also Dupoux, 2018; Räsänen, 2012). ***The learner model*** must be described in terms of the high-level computations it performs on the sensory input, and these high-level principles must be translated into specific learning algorithms that operate on a computer and can deal with real acoustic speech input (cf. Marr, 1982). In addition, any simulation of the learning process requires the specification of ***an environment model*** (de Seyssel et al., 2023). At minimum, this involves defining the speech input to which the learner is exposed, but it may also incorporate multimodality, interactivity, or other factors relevant to the study. Finally, any computational study requires a protocol for comparing the model's learning outcomes either against data from real infants or against a linguistic ground truth describing what the listener should decode from the input—often referred to as ***the outcome model*** (de Seyssel et al., 2023). The overall ecological plausibility and validity of any computational learning simulation depend on all three of these components (Fig. 1).

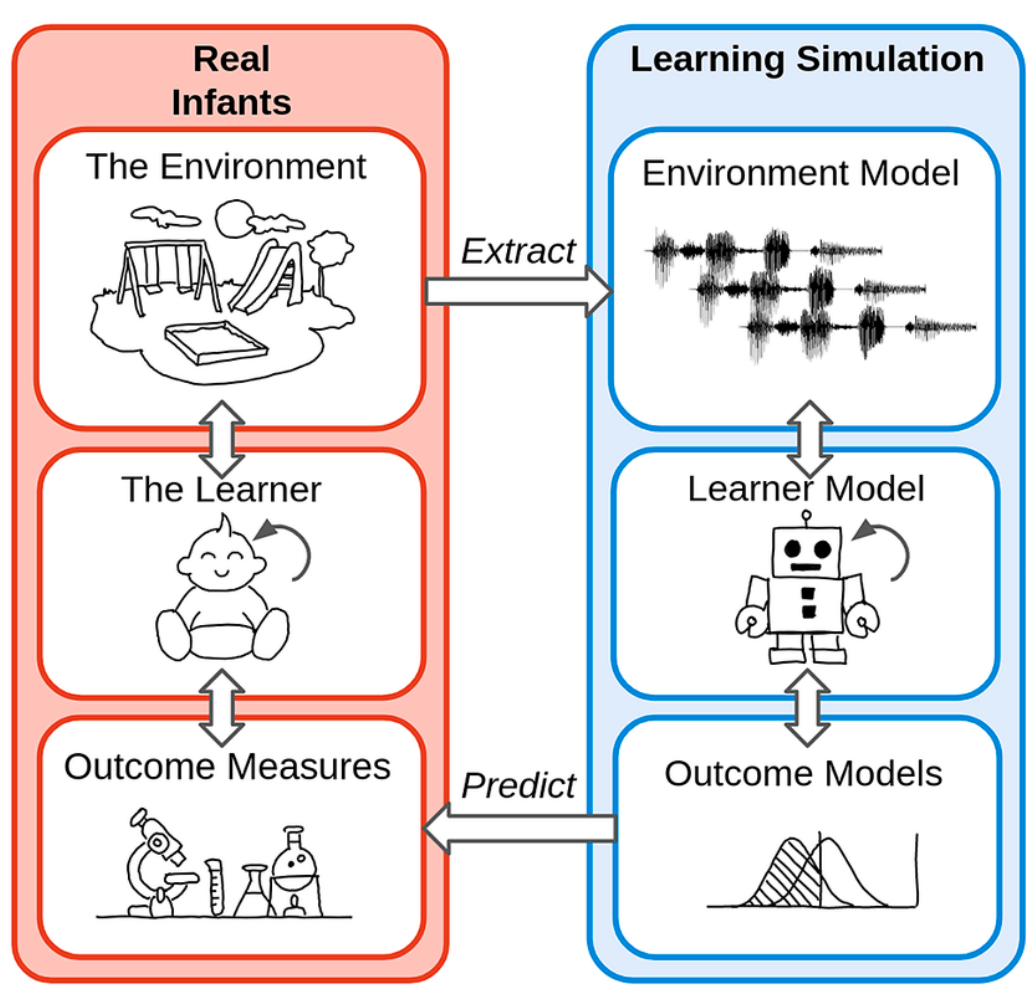

**Figure 1**: Illustration of the key components in computational modeling of early language learning: the environment model, the learner model, and outcome model. Reproduced from de Seyssel et al. (2023) under the CC-BY license.

Because of their explicit nature, computational models serve directly as mechanistic theories of the learning process. In addition, their integrative nature allows computational models to study multiple aspects of language in parallel. Since simulations are not restricted by real-time considerations, language learning can be modelled across the developmental timeline, under varying environmental conditions (e.g., different language inputs), and without concerns about participant fatigue or fussiness when testing models across a large number of test trials or broad range of linguistic capabilities. This means that computational models of early learning can act as an integrative force, supporting hypothesis testing and theory formation for increasingly holistic models of early language development. They also provide learnability proofs and generate new hypotheses about how language skills might emerge from the learning conditions available to real infants. In doing so, these models address the current need for more comprehensive and accurate theoretical accounts of early language development (Rowland et al., 2026).

The rest of this chapter focuses on reviewing several recent developments in computational modeling of the very first stages of language development from speech or multimodal input. The focus will be on models that attempt to learn directly from real acoustic speech, as one of the great challenges in early language learning is how to bootstrap the basic speech comprehension skills from continuous and variable speech input. In contrast, models operating on discrete inputs (e.g., Frank et al., 2010; Warstadt et al., 2023) are not discussed. This is because there are currently no mechanistic explanations for how infants might acquire categorical (invariant) speech representations, nor is it clear whether such representations are even involved in early language (see, e.g., McMurray, 2022, 2023; see also Schatz et al., 2021). Similarly, models operating on other types of abstract speech features are outside the current scope (e.g., Bhat et al., 2022), and their link to acoustic models of learning remains to be established.

## Modeling language acquisition with modern machine learning

As motivated above, learning a language from speech input is far from trivial. In the absence of an existing system of linguistic units, such as phonemes or words, the learner has to begin by analyzing acoustic waveforms for statistical structure, even though nothing repeats in exactly the same way. To enable computational models to handle the complexity of real-world speech data, digital signal processing (DSP) and machine learning (ML) are typically used in models of early language learning, with an increasing emphasis on the latter in modern approaches. DSP is commonly used to handle processing aspects that are assumed to be physiologically hard-wired (e.g., time–frequency analysis in the peripheral auditory system) or otherwise innate and not subject to learning (e.g., loudness-based chunking of input into syllable-like units; Räsänen et al., 2018). It is also used to apply predefined transformations for data preprocessing (e.g., mitigating differences in audio recording setups) or to probe what the model has learned. Since ML is all about identifying statistical patterns in data, it is used to implement different forms of statistical learning mechanisms. Note that the goal is not to simulate a child or the environment in their entirety, but to specify the minimum set of information processing principles and environmental factors that result in human-like language learning outcomes.

Early computational models of language learning typically employed one or more dedicated processing mechanisms to solve specific sub-problems of the learning process. For instance, one category includes models of speech segmentation, where various computational solutions for unsupervised segmentation of phones (e.g., Michel et al., 2017; Scharenborg et al., 2007), syllables (e.g., Räsänen et al., 2018), and words or recurring phrases (e.g., Elsner & Shain, 2017; Park & Glass, 2006) from running speech have been investigated. Another set of studies has investigated the categorization of speech sounds into phonetic or syllabic categories using different clustering strategies with acoustic (e.g., de Boer & Kuhl, 2003; Seshadri et al., 2017; Vallabha et al., 2007; see also Matusevych et al., 2020) or audiovisual (e.g., Coen, 2006) input.

In addition, models exploring synergies across multiple levels of linguistic structure have been proposed across the years, such as joint learning of phonemes and words (Feldman et al., 2013), segmentation and categorization (Kamper et al., 2016), or learning from speech and concurrent (simplified) visual input (Räsänen & Rasilo, 2015; ten Bosch et al., 2009). These models demonstrate how interactions between different levels of language—and between language and the external world—can support the bootstrapping of early language comprehension skills rather than merely complicating the learning process (see also Johnson et al., 2010, for discussion). Moreover, some of these findings have contributed to core theoretical questions in spoken language research, such as whether phonemic categories are necessary to explain empirical data on phonemic discrimination (see Schatz et al., 2021) and whether word segmentation as a separate mechanism is required to bootstrap lexical learning (Räsänen & Rasilo, 2015).

Over time, modeling research has gradually shifted from sub-problem-specific models toward more integrated learning models with increasingly realistic representations of the learning environment. A central driver of this progress has been the series of Zero Resource Speech (Zerospeech) Challenges, organized since 2015 (Dunbar et al., 2022; see also Jansen et al., 2013, for early origins). The aim of these challenges has been to develop unsupervised learning algorithms that could learn different types of linguistic patterns from acoustic speech without labeled data (hence referred to as a “zero-resource” problem). These challenges formalized and standardized evaluation practices for several aspects of language learning, such as measuring models’ phonemic discrimination capabilities with ABX testing, in which a model’s internal representations for phonemic segments are compared within and across phonemic categories to understand whether the model internally represents phonemic contrasts (Schatz et al., 2013). Other evaluations have included, e.g., scoring word-segmentation accuracy, benchmarking models on lexical and syntactic acceptability judgments, comparing semantic similarity judgments of word pairs with those of humans, and, more recently, evaluating prosodic parsing skills (see www.zerospeech.com). At the same time, the Zerospeech Challenges have encouraged the development of models that can learn in similar ways across a range of languages and capture multiple aspects of language simultaneously.

### 2.1 Learning through prediction

Recently, a new family of ML techniques known as self-supervised learning (SSL) has emerged for representation learning without external supervision. These models, implemented as deep neural networks, can learn useful general-purpose representations of input data without external labels. In practice, such models are typically used for various downstream tasks in technological applications, such as automatic speech recognition, speaker diarization, or speech emotion analysis (e.g., Yang et al., 2021), when paired with a small amount of labeled data for the task of interest.

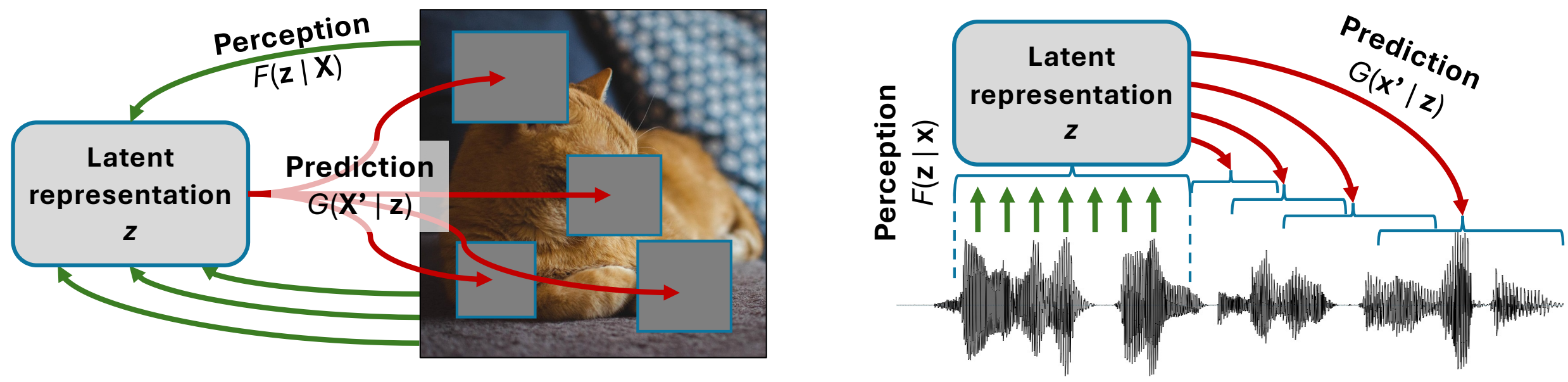

**Figure 2:** A schematic view of generic self-supervised prediction tasks, where unobserved image contents are predicted from the spatial context (left) or future speech input is predicted from the preceding context (right). In the figure, **z** stands for internal (latent) representations, *F*(**z** | **X**) for a perception process that maps incoming sensory input **X** into a latent representation **z**, and *G*(**X’** | **z**) stands for a predictive process that tries to predict future or otherwise unobserved sensory input **X**, based on the current latent state **z**. Note that in practice, predictions could also concern lower-level representations of the input instead of predicting the sensory signals themselves.

What makes SSL particularly interesting for modeling cognitive phenomena is that its learning process does not directly target any specific representations of the input. Instead, learning is achieved by setting up some kind of data-driven proxy task for the neural network to solve, such as predicting masked sections of images or the future continuation of a speech signal (Fig. 2). If the task is sufficiently challenging, the model must learn internal representations that effectively support the prediction of unseen content. During training, these models access partially hidden data, predict the unseen parts, then observe the true input and update their network weights to reduce future prediction errors in similar situations. By repeating this process over hours of speech or thousands of images, the network gradually learns latent structural representations of the data that support “filling in the blanks.”

For example, to predict masked sections of real-world photographs, one cannot simply extrapolate from the nearest pixels into the missing area. Instead, a general understanding of the overall scene is required to infer what could be missing from the picture, and this must then be translated into detailed predictions of the visual characteristics of the missing elements (see Fig. 2, left). In the same vein, to predict the future speech signal from the past, the model might need to learn about possible speech sounds and their probabilistic sequential structure (i.e., phonotactics), while knowledge of words and word-to-word dependencies (syntax) would further improve prediction accuracy. This perception–prediction mechanism shares similarities with human infants, who have access to continuously unfolding auditory input or fixation-based partial sampling of the visual environment, allowing continuous prediction and comparison with actual input.

Importantly, the latent representations that an SSL model learns are not directly determined by the researcher. Rather, they emerge from the conditions set by input data, the learning algorithm (including the proxy goal and the method of parameter updating), the neural architecture, and the initial parameter values of the network. SSL models can also learn from various data modalities with similar core processing principles and can operate in a temporally causal manner. This means that SSL and the closely related techniques for cross-modal associative learning can be applied to several concurrent sensory modalities and motor outputs, coupling the latent states of different modalities through cross-modal predictions. In other words, internal states induced by one sensory stream (e.g., hearing) can be used to predict internal states of another sensory stream (e.g., vision) and vice versa. Overall, this enables an array of predictive tasks for both human and computational learners, as illustrated in Fig. 3.

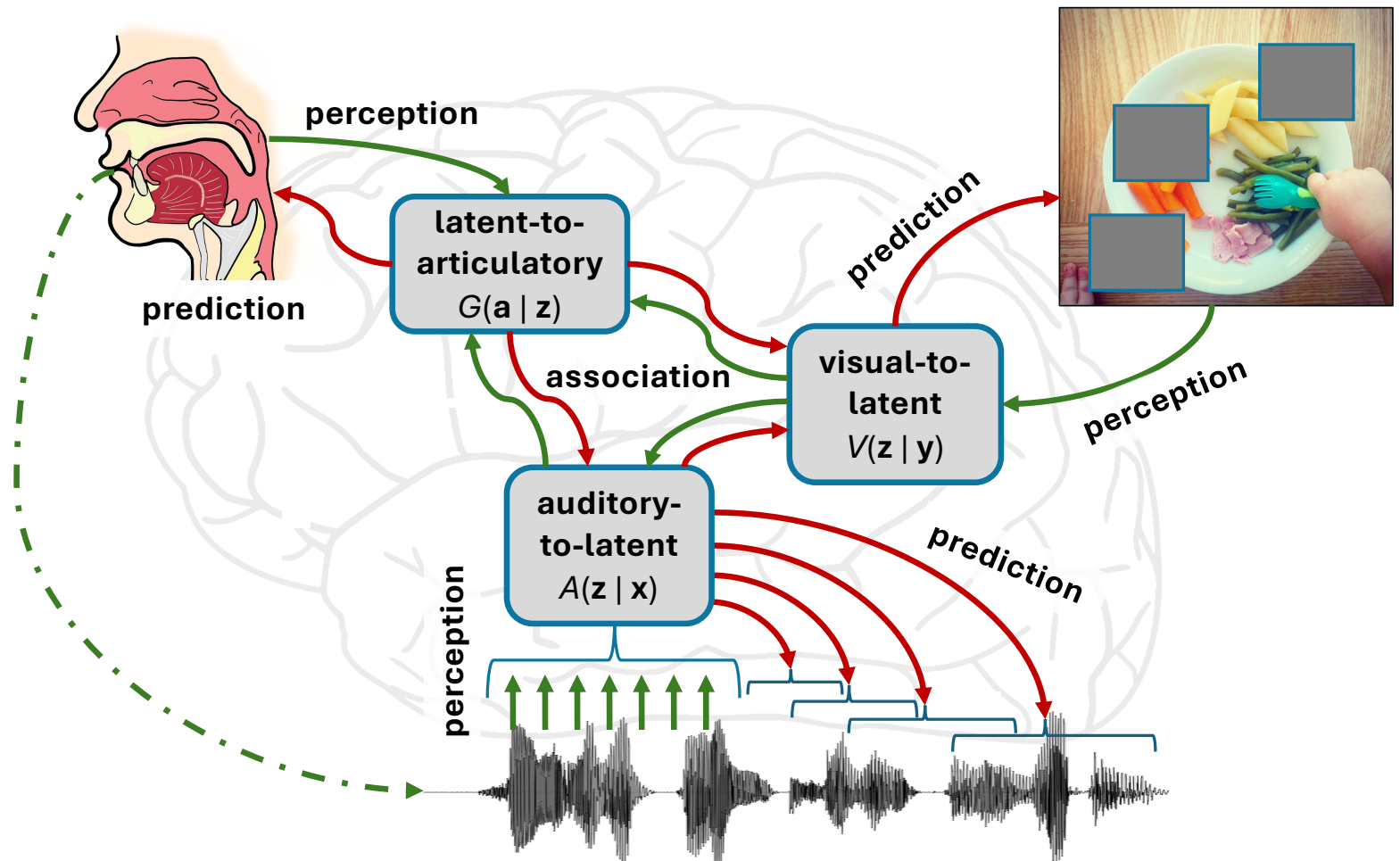

**Figure 3:** A schematic view of some of the key predictive affordances for a hearing and seeing language learner. Perception extracts internal (latent) representations **z** of the sensorimotor world, such as auditory input **x** and visual input **y**. These latent representations can be used to predict temporal and cross-modal unfolding of further experiences, including also articulatory control parameters **a** as predictions (cf. Friston, 2010). Learning can be guided by minimization of the respective prediction errors. Relation of functions to real cortical areas is purely illustrational.

At a high level, SSL models are also linked to theories of human cognition, including the idea of the human brain as a predictive machine characterized by constant interaction between top-down predictions and bottom-up sensory processing (e.g., Friston, 2010; Rao & Ballard, 1999), and to constructivist views of development, where existing internal representations guide incremental assimilation of new information and future learning (Rowland et al., 2026). Although mainstream SSL models do not explicitly implement specific cognitive or neuroscientific theories, they share the same high-level core principles: sensory input influencing the system's internal states across multiple levels of representational hierarchy; internal states generating predictions of new sensory input; processing of incoming input through comparison with those predictions; and learning occurring as a consequence of mismatches between predictions and real input.

Overall, all the above factors make SSL-based computational models of statistical learning attractive, ecologically relatively plausible, and theoretically parsimonious (but see the end of the chapter for a discussion of limitations). In the context of early language development, this leads to the question: can we simulate early language acquisition with SSL-based computational models that implement one or more of the predictive affordances illustrated in Fig. 3? More specifically, can we explain early language as an emergent latent representational system that supports effective intra- and cross-modal predictions in the sensorimotor domain, rather than as a process in which the learner directly attempts to acquire specific aspects or units of language? Is it possible to learn phonemes and words simply by predicting future audio over time? Can such models simultaneously learn multiple aspects of language with one generic predictive mechanism? How does concurrent visual input help or hinder the process? Do SSL models require specific a priori biases or constraints to succeed in learning language patterns?

The idea of language representations being an emergent by-product of predictive optimization, rather than a direct target of learning, is formulated as the Latent Language Hypothesis in Khorrami and Räsänen (2021). The idea has its roots in earlier works, such as Elman's (1990) insight that syntactic structures can emerge from temporal predictions of linguistic input, and in the dynamical systems approach of Thelen and Smith (1994), where connections between perception, action, and multimodality form the basis for the self-organization of categorical representations. The hypothesis is also a natural consequence of the general predictive brain framework, in which internal latent states of the brain act as explanatory—that is, predictive—causes for externally observed sensory stimuli, and in which learning and action serve to reduce the divergence between internally generated predictions and actual sensory stimulation (see, e.g., Friston, 2010).

To date, several modeling studies have investigated early language learning using SSL-based learners, and the rest of this chapter reviews these developments. We start by introducing SSL-based models that learn from acoustic speech input. We then turn to audiovisual models of learning, exploring how multimodality may help bootstrap the early language system. Before concluding, we discuss recent advances toward increasingly realistic learning simulations, including progress in modeling the learning environment and in evaluating models' language skills in comparison with those of real infants.

## Learning from raw audio with self-supervised learning

One predictive affordance available to learners with a functional hearing system is to predict the evolution of the speech signal over time, given access to past speech input up to the current moment. This prediction task can be implemented with various neural network architectures. The main requirement is that the network should be able to perceive and represent a sufficiently long history of past speech input, and then use this "context representation" to predict the future signal[2].

Temporal prediction can be implemented in two basic ways: (1) predicting the future low-level acoustic signal, or (2) predicting the model's own internal representations corresponding to the future signal. As a coarse analogy to the biological brain, one could think of these as predicting patterns of auditory nerve stimulation in a top-down manner or predicting stimulus-driven activity in higher-order cortical areas (e.g., belt or parabelt regions of the auditory cortex). While the former are primarily driven by physiologically hard-wired processes, the latter are subject to experience-based plasticity and learning. However, it is important to note that the computational models discussed here are not intended as neurophysiological but algorithmic models of the learning process.

At a technical level, an example of the first class of models is the autoregressive predictive coding (APC) model (Chung et al., 2019), illustrated in Fig. 4 (left). The model takes as input a logarithmic Mel spectrum—a spectral envelope representation of the speech signal that mimics the frequency resolution of human hearing—and processes the sequence of log-Mel feature frames (multivariate vectors) appearing every 10 milliseconds up to the current time through a set of recurrent neural network (RNN) layers. The output of the last RNN layer is then fed into a linear layer that projects the current latent activation vector (context vector) into a prediction of a future log-Mel vector $k$ steps ahead in time. The predicted and actual log-Mel vectors are compared using the Euclidean or Manhattan distance, and the resulting prediction error is backpropagated through the network to update its parameters. As shown in the original study by Chung et al. (2019), training the APC model on large amounts of speech data results in hidden-layer representations that encode both phonetic structure and speaker identity in the speech input, making these information sources easily separable (see also Liu et al., 2023, for further analysis).

APC is conceptually very similar to the classical Elman (1990) RNN model. Where Elman's model was tasked with predicting discrete letters and words, APC learns directly from real speech and thus leads to different types of linguistic representations emerging within the two models—lexical categories and type/token distinctions in the former, versus segmental and suprasegmental information in the latter. Moreover, modern large language models are based on the same predictive principle (e.g., Radford et al., 2019), though they operate on textual representations.

[2] Prediction of masked speech segments is another strategy for speech-based SSL, which can simply be viewed as a non-causal version of the future prediction task.

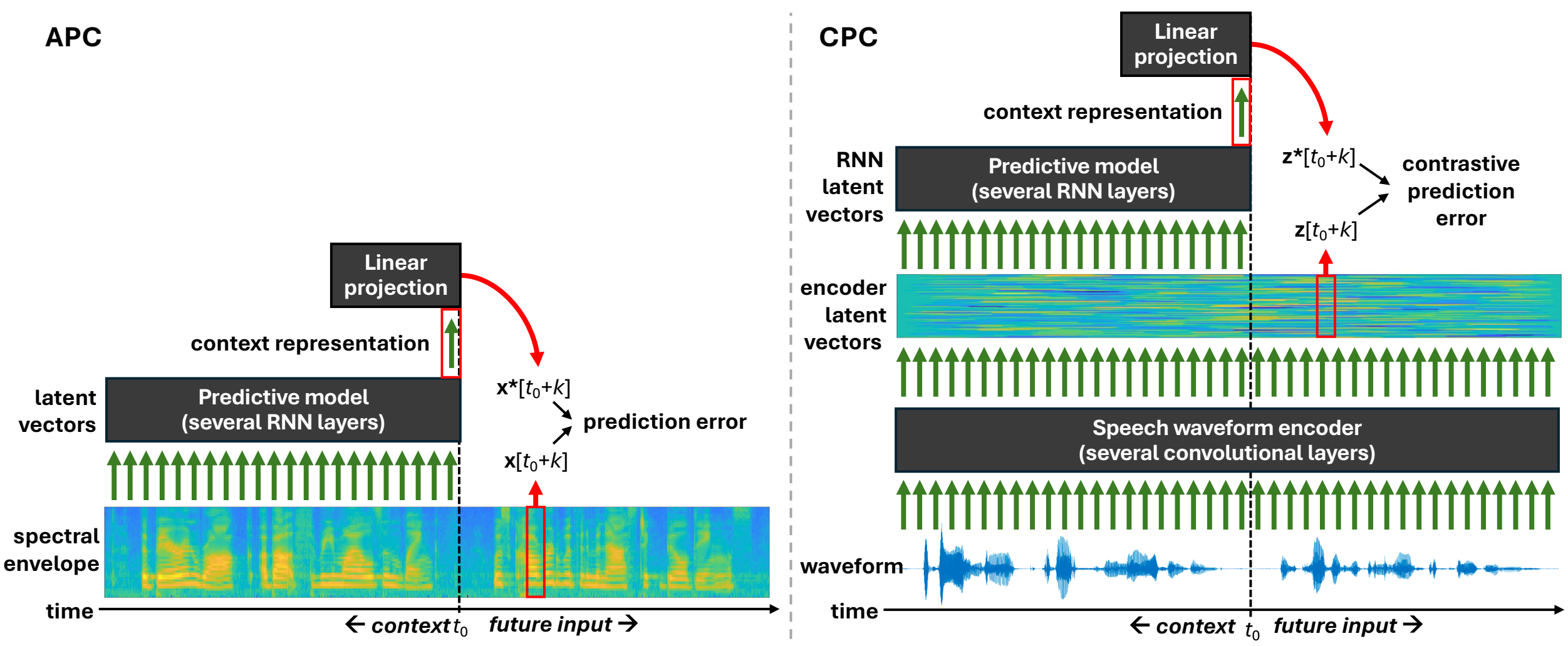

**Figure 4**: Autoregressive predictive coding (APC) (left) and Contrastive predictive coding (CPC) (right) as examples of SSL-based learning algorithms. Adapted from Bäckström et al. (2022).

However, while APC is already a powerful learning algorithm, the task of predicting the full spectral detail of the speech signal explicitly discourages the model from developing internal representations that abstract away from low-level variability of the signal. Contrastive predictive coding (CPC; van den Oord et al., 2018) relaxes this assumption by allowing the network to invent its own prediction targets (Fig. 4, right). This is achieved by dividing the model into two components: (1) a non-linear encoder that maps the raw acoustic speech waveform into latent representation vectors in short time windows (e.g., one vector every 10 ms), and (2) an autoregressive module, similar to that in APC, that accumulates past latent observations into a context representation and predicts the future. Instead of predicting the low-level signal, the context vectors from the autoregressive module are linearly mapped into predictions of future encoder outputs. Hence, CPC simultaneously learns what to predict and how to predict it.

To prevent the neural network from learning a trivial solution—for example, representing and predicting all speech with a constant-valued vector—contrastive loss is used in CPC training. Instead of minimizing the difference between predicted and real future observations, the network learns representations and a predictor that allow it to distinguish real futures from plausible but incorrect ones drawn from other moments in the input speech stream. As a consequence, CPC is forced to invent an internal code for the speech input that enables it to separate likely true futures from unlikely ones (which may still be likely in other temporal contexts). However, it does not have to encode and predict all variability present in the low-level signal. For instance, the model might learn internal representations for elementary speech sounds (phones) to exploit phonotactic regularities. At the same time, its representations may become more invariant to pronunciation variability or background noise if these factors do not contain systematic cues relevant to the prediction task. In the original paper, van den Oord et al. (2018) showed that training CPC also results in latent representations that effectively and simultaneously encode both phonetic and speaker information, which are linearly separable within the representation space.

Overall, the speech representations learned by APC and CPC differ clearly from classical low-level signal features. In classical features, such as logarithmic Mel features or Mel-frequency cepstral coefficients (MFCCs, Davis & Mermelstein, 1980), the various types of information and variability carried by the speech signal are much more intertwined in a complex and largely non-linear manner. In contrast, CPC and APC can resolve distinct systematic aspects of the speech data in a purely unsupervised manner, mapping the waveform or spectral envelope into a high-dimensional space with separate subspaces for different types of signal content.

**Using APC and CPC as models of early statistical learning in infants**

Given the capability of APC and CPC to learn patterns from speech input in an unsupervised manner using generic predictive mechanisms, language development researchers have begun investigating their feasibility as models of early phonetic and lexical learning. The original papers by Chung et al. (2019) and van den Oord et al. (2018) demonstrated that phonetic learning is possible using APC and CPC. However, these studies used fixed datasets of 100 hours (CPC) and 360 hours (APC) of speech for model training, and phonetic knowledge was evaluated using a linear classifier trained to predict phonetic labels for each short-term latent representation of speech.

To better understand learning in these models relative to real infants, Lavechin et al. (2025) explored phonetic and lexical learning in CPC as a function of the developmental timeline of a simulated infant. They used a CPC model variant referred to as STELA (STatistical Learning of Early Language Acquisition). In this model, k-means clustering is applied to the learned internal representations of CPC to convert them into discrete categories of speech events, substituting each multivariate latent vector with an integer token denoting the nearest cluster centroid index. Using the resulting discrete token sequences of speech data as input, the authors then trained a recurrent neural network for next token prediction—an analogical task to how standard language models (LMs) are trained to predict tokens for letters, words, or word-pieces based on the preceding context. This predictor for internal representations can be then used to evaluate acceptability of different speech inputs by comparing the probabilities that the predictive process assigns for different stimuli, such as words versus non-words (cf., transition probabilities of syllables often discussed in statistical learning). Their learning experiments were conducted in both English and French, enabling evaluation of both native and non-native phonemic and lexical discrimination skills of the model.

For phonemic discrimination, the ABX test of Schatz et al. (2013) was used. The purpose of the ABX test is to assess whether the model can represent tokens of the same phoneme with similar activation patterns while treating tokens from different phoneme categories as distinct, thereby reflecting the phonemic discrimination capability of the learner. The idea is analogous to how infants' phonemic discrimination is tested in the lab, where preferential listening paradigms are used to probe whether the listener treats two stimuli as the same or different. In the ABX test, this is achieved by feeding the computational model speech samples containing two contrastive triphone segments, A and B (e.g., a minimal pair A = "*bat*" and B = "*bit*"), and extracting the corresponding latent representations. A third input X, containing the same triphone as A but extracted from another speech signal, is then fed to the model, and the distances *d*(A,X) and *d*(B,X) between the latent representation of X and those of the two reference tokens are calculated. If the distance between X and A is smaller than the distance to the across-category exemplar, the trial is considered successful discrimination (i.e., $d(A,X) < d(B,X)$); otherwise, the trial is considered failed. By subjecting the model to thousands of contrastive pairs extracted from a speech corpus with phonetic annotation, the phoneme discrimination rate (ABX accuracy) can be calculated as the percentage of successful trials.

For lexical acceptability, a spot-the-word task was applied. The task is based on a test originally proposed by Le Godais et al. (2017) for text-based models. It involves feeding the model minimal pairs of isolated real words and nonwords from the language's lexicon (e.g., "camp" vs. "pamp") and comparing the probabilities the model assigns to each item. If the model considers the real word more likely than the non-word, the trial is considered successful. By computing the proportion of successful trials across a large number of word pairs, an overall lexical discrimination score is derived. To adapt the task to their simulations, Lavechin et al. identified the words present in their model's training data (the simulated environment) and constructed minimal pairs for them, ensuring that both words and nonwords matched in syllabic and phonotactic structure. The words were then synthesized into speech stimuli in different speaker voices using a speech synthesizer (Google Text-to-Speech API).

To train the model, Lavechin et al. (2025) exposed it to speech input ranging from 50 to 3,200 hours per language (English or French) to simulate the amount of linguistic experience accumulated by infants up to three years of age (Cristia, 2023; see also Coffey et al., 2024, for infant speech input estimation). The speech material was extracted from English and French audiobooks covering a range of text genres. After training STELA for a given number of speech hours in a given language, the model was then tested on its performance in the ABX and spot-the-word tasks in both the training language (native) and the other language (non-native). This approach provided developmental trajectories for the two language capabilities as a function of the learner's simulated speech experience, separately for native and non-native languages.

As a result, the authors found that STELA was successful in phonemic learning, reaching a native-language ABX score of over 81% after 3,200 hours of speech and already achieving 75% discrimination accuracy after just 50 hours of input. Notably, the native-language ABX score was significantly higher than that of the non-native language throughout the developmental trajectory. This demonstrates that the model successfully replicated the native language bias in phonemic discrimination, a well-known phenomenon in early language development (e.g., Kuhl et al., 2006; Werker & Tees, 1984). However, the results also showed a steady increase in non-native discrimination with learning rather than the decline observed in infants. Yet, as Lavechin et al. note, the study tested discrimination across all possible contrasts in the non-native language, whereas empirical data on infant non-native discrimination are largely based on a carefully selected set of contrasts that exist only in the non-native but not in the native language. These differences make direct comparison between the model and infants challenging.

As for lexical learning, the native language lexical score reached almost 63% after exposure to 3,200 hours of speech input. While this performance is substantially above chance level (50%), the model could distinguish only some words from their non-word counterparts at the simulated age of three years. The relatively slow rate of further improvement with additional input suggests that lexical learning in STELA remains incomplete compared with learning in real infants. Still, the result shows that statistical learning operating on raw acoustic speech input can bootstrap lexical learning, at least to some extent. Moreover, closer analyses by Lavechin et al. (2025) revealed that individual words with high lexical scores were also associated with higher frequency of occurrence in the input data, replicating a well-documented phenomenon in early language (e.g., Ambridge et al., 2015).

Overall, the results from the study by Lavechin et al. (2025) demonstrate that statistical learning operating on real speech can successfully bootstrap phonemic and lexical learning from input data on a realistic scale. Notably, this learning occurs without any built-in linguistic priors or other strong innate constraints and instead relies solely on predicting the future evolution of the signal from past context.

As for other studies, Cruz Blandón and Räsänen (2020) showed that APC and CPC can learn phonemic knowledge from only a few hours of speech in Mandarin and French. Recently, Liu et al. (2024) used a CPC-based SSL learner to investigate temporal dynamics of phonetic representations in the model. They found that the model was able to replicate temporal characteristics and contextual invariance of phonetic representations observed in the brain during a speech perception task, demonstrating how a link between SSL models and neurological data can be established.

## Learning from audiovisual input

While it is obvious that the development of spoken language is driven by access to speech input, everyday speech communication does not occur in isolation. Indeed, much of the caregiver speech in infant–caregiver interaction relates to ongoing activities in everyday contexts, providing the basis for grounding words in their referential meanings. For instance, early child vocabularies are populated by nouns and verbs that have concrete correspondences in visual experience (e.g., the Communicative Development Inventories, CDI, Fenson et al., 2007; and large-scale statistics in Wordbank, Frank et al., 2017). More generally, there appears to be a high degree of systematicity in what infants hear and see in everyday contexts (e.g., Clerkin & Smith, 2022).

Infants' access to multimodal experiences with the world suggests that computational models of early language development should likewise take multimodality into account. While multimodality is required for grounding speech in the "things" out there in the world (cf. Quine, 1960), visual input correlated with auditory speech may also support the parsing of continuous speech input into meaningful patterns that have multimodal predictive value, such as words or phrases, thereby bypassing the need for explicit segmentation mechanisms in perception (Räsänen & Rasilo, 2015). Within the predictive processing and statistical learning frameworks, this would correspond to minimization of predictive uncertainty across sensory modalities (see Fig. 3): by associating heard speech with seen objects and events through tracking cross-situational statistics and thereby overcoming referential ambiguity (cf. Smith & Yu, 2008), the learner could bootstrap their language system through holistic mappings between speech form and meaning. With increasing speech experience, the learner could then gradually work toward compositional and more invariant representations of language.

Learning to associate acoustic speech with visual input would align with usage-based theories of language (Tomasello, 2000; see also Merkx, 2022) and would also be consistent with the PRIMIR framework by Werker and Curtin (2005). In PRIMIR, phonemic knowledge is hypothesized to gradually emerge from an early lexicon consisting of acoustic word forms and their meanings, which are initially encoded with general (non-linguistic) perceptual representations (see also Curtin et al., 2011). However, the main challenge in audiovisual learning is the substantial referential ambiguity that exists between heard speech and observable visual scenes (Quine, 1960), where usually only some words in larger utterances refer to some specific aspects of the visual scene. To directly learn from audiovisual input, the learner must identify relevant segments of the auditory stream (words) and relevant objects of the visual scene (referents), and then track which words go with which referents (Fig. 5).

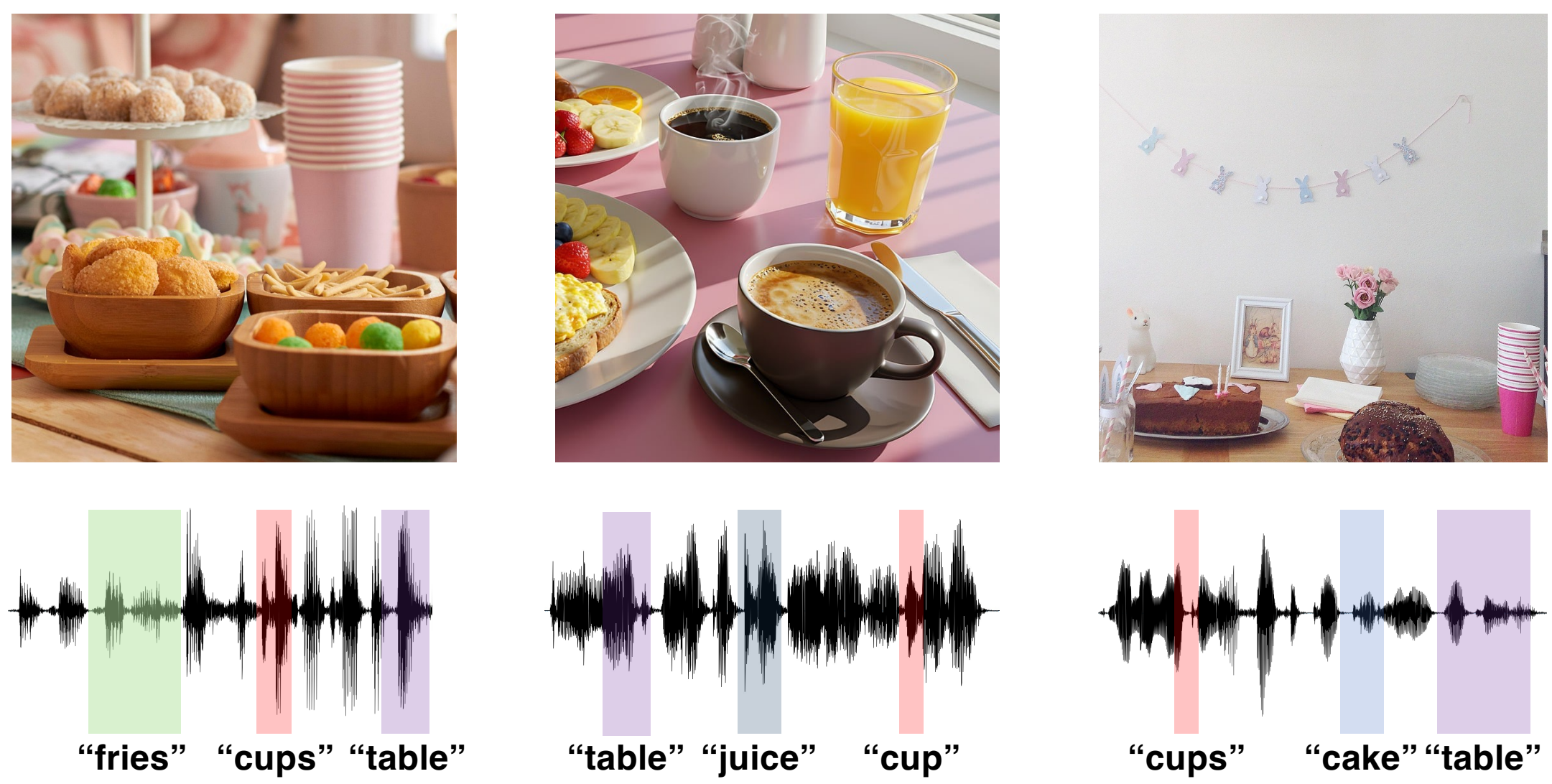

**Figure 5**: An illustration of referential ambiguity in cross-situational audiovisual learning, where words almost always occur as parts of longer utterances, and where many potential visual referents exist in each learning situation. Adapted from Khorrami and Räsänen (2025).

To study visual grounding in speech learning, a series of computational models known as visually grounded speech (VGS) models have been proposed in recent years. The basic idea behind deep neural VGS models is that they consist of three main components: a speech encoder responsible for processing the acoustic speech stream, a visual encoder that processes pixel-level images, and an associative mechanism that maps the outputs of the two modality-specific encoders into a shared representation space (Fig. 6; Harwath et al., 2016; see also Chrupała, 2022, for a review).

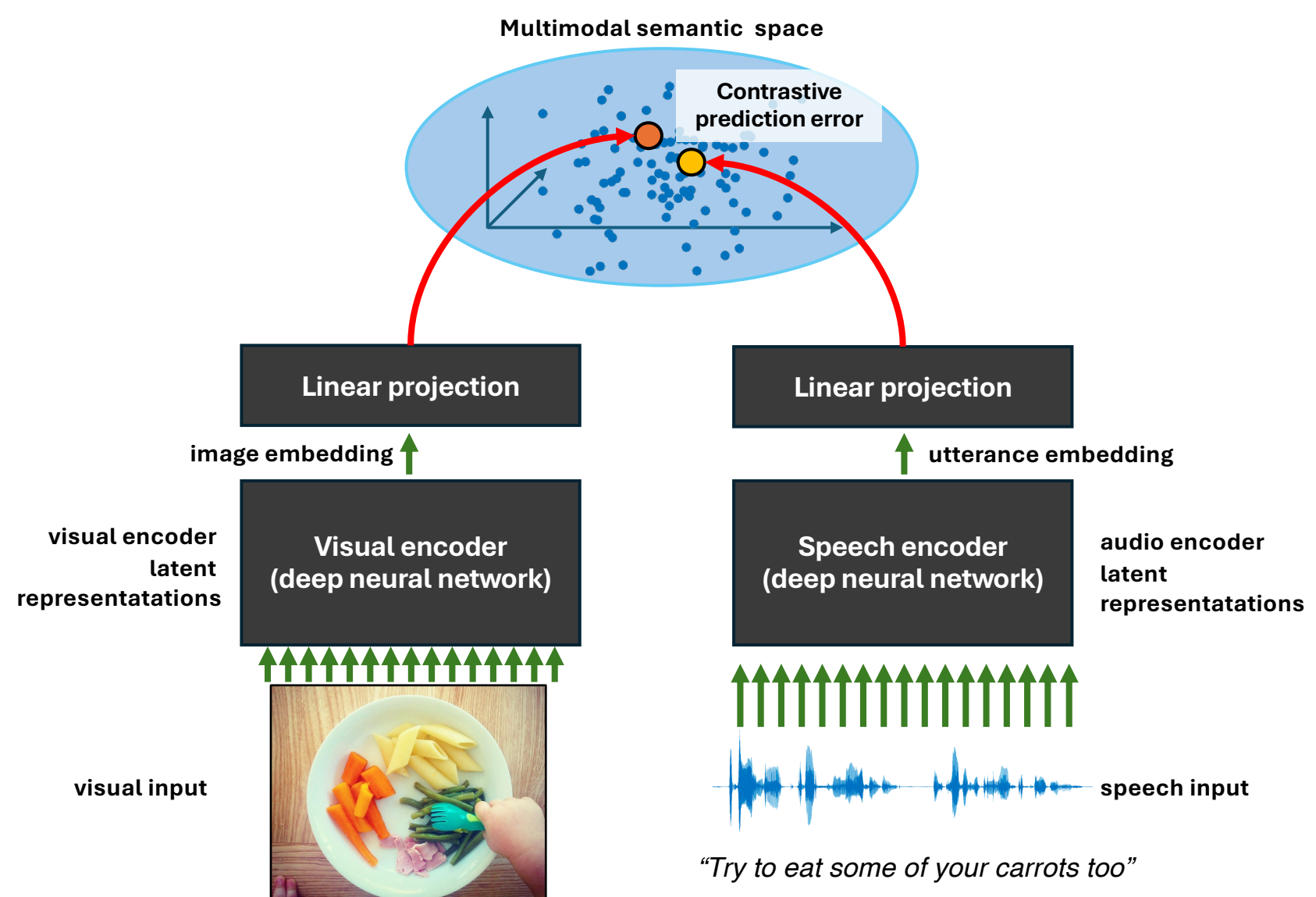

**Figure 6:** A schematic view of visually grounded speech models. Modality-specific encoders are responsible for extracting representations that support cross-modal matching of information between the sensory streams. Adapted from Harwath et al. (2016) and Khorrami and Räsänen (2021).

During training, the network is presented with an image and a concurrent speech utterance describing it, producing feedforward activations in the two encoders. The audiovisual associative module receives the activations ("embeddings") from both encoders, computes their similarity, and updates the associative mappings and encoder layers to increase their future similarity. To avoid a collapse of representations into a trivial solution, this objective is implemented as a contrastive loss similar to CPC (see also Gutmann & Hyvärinen, 2010) or as a margin-based loss (Schroff et al., 2015). In both cases, the model must learn to discriminate true matching (concurrent) inputs from false but otherwise possible input pairs, and where the two alternative losses only differ in their mathematical details of how the process is implemented in practice. As learning progresses, the outputs of the final (deepest) layers of the modality-specific encoders (often referred to as embeddings) become increasingly similar when semantically related input is observed in both. To support "prediction" or "translation" between the modalities, layers of the modality-specific encoders must learn representations that encode information relevant to the task. Hence, the pressure to learn useful speech representations arises from the need to extract patterns that have maximal predictive value with respect to internal representations co-existing outside the auditory domain.

### Language learning simulations with VGS models

In the original paper proposing the basic modern VGS architecture for speech, Harwath et al. (2016) did not explicitly study the model from the perspective of an infant language learner. Instead, the main evaluation task focused on speech-based image retrieval. By training the model on 111,400 utterances paired with corresponding photographs of everyday scenes, the model learned to associate spoken descriptions with semantically similar images. However, Harwath et al. also conducted a set of qualitative post hoc analyses of what the model had learned. These revealed that the temporal activations of the speech encoder were highly correlated with the image encoder outputs during words that referred to entities visible in the corresponding images. They also observed qualitatively that neural activation patterns of different tokens of the same word tended to cluster well in the representation space, indicating that the model had learned to discriminate auditory word forms from one another.

In a study by Chrupała et al. (2017), the authors used a speech synthesizer to create five spoken versions of written captions for over 300,000 everyday-scene photographs from the MS COCO database (Lin et al., 2014) and then trained a VGS model using the resulting speech–image pairs. In contrast to Harwath et al. (2016), who used convolutional neural networks (CNNs) for the speech encoder, Chrupała et al. employed a recurrent highway network (RHN)-based encoder architecture; a variant of RNNs (i.e., networks that maintain a representation of the past context in a dedicated hidden state that then contributes to the processing of the current input). They then systematically analyzed the model's hidden layers for their ability to (i) predict utterance length, (ii) detect individual words in the input, (iii) capture sentence-level semantic similarity ratings in relation to corresponding human data, and (iv) perform homonym disambiguation (distinguishing words with different spellings but identical pronunciations).

The results of Chrupała et al. showed that the hidden RHN layers indeed contained information about utterance length and the presence of target words in the input utterances. They also encoded the semantic relatedness of sentence pairs with correlations above r = 0.6 relative to human ratings, and homonym disambiguation improved systematically from early to deeper layers of the network. Overall, access to lexical and semantic information appeared to improve toward the deeper layers of the speech encoder (see Chrupała et al., 2017, for details).

In Alishahi et al. (2017), the authors conducted a systematic follow-up analysis examining what kinds of phonological information the VGS model from Chrupała et al. (2017) learned to encode from the speech input. Their analysis showed that phonemic information was most accessible in the early layers of the RHN encoder, achieving over 90% ABX accuracy for vowels and substantially above-chance performance for various consonant types. They also conducted an agglomerative clustering analysis of neural activations and observed that phonetic features with similar manners of articulation tended to share similar activation patterns, with some confusions among similar places of articulation. Overall, the results were qualitatively similar to EEG analyses of human auditory responses to speech (Khalighinejad et al., 2017). Finally, they showed that the model's early and middle layers were able to perform synonym discrimination accurately, indicating that the corresponding representations were more phonological than semantic in nature. The ability for synonym disambiguation, however, disappeared in the final layers of the network before multimodal fusion, showing how the mapping from form to visual semantics progressively abstracts away from phonological details (see Alishahi et al., 2017, for details).

Havard et al. (2019a) analyzed which segments of the speech input a VGS model attends to when solving the audiovisual grounding task, using both Japanese and English data. They found that a model trained on English primarily attended to nouns in the speech input, whereas determiners, adjectives, and adpositions were rarely attended to. In contrast, in addition to nouns, the Japanese model attended to particles, especially the so-called GA particle that marks the preceding word as the subject of the sentence—likely a key aspect of visual scene descriptions. Hence, the model had learned in both languages to attend to linguistic content that was helpful for disambiguating alternative visual scenes. This pattern aligns with the noun bias hypothesized for early child language (Gentner, 1982) and with findings on Japanese infants using GA particles for speech segmentation before other particles (Haryu & Kajikawa, 2016).

Since most existing studies used synthetic speech (e.g., Alishahi et al., 2017; Chrupała et al., 2017; Havard et al., 2019a, 2019b; but see Harwath et al., 2016), it was unclear whether analyses conducted on synthetic input were representative of real speech. In addition, some studies used CNN-based speech encoder architectures (Harwath et al., 2016), while others used RNN-based models (Alishahi et al., 2017; Chrupała et al., 2017; Havard et al., 2019), leaving open how much the learning outcomes depend on the chosen neural architecture. To address these shortcomings, Khorrami and Räsänen (2021) investigated phonemic, syllabic, and lexical learning in CNN- and RNN-based VGS models trained on both synthetic and real speech input. In addition to examining whether linguistic unit identities could be decoded from the model layers, they also evaluated the models' ability to implicitly segment the speech input into these units by inspecting the temporal patterns of hidden-layer activations.

The results of Khorrami and Räsänen showed that (i) phonemic, syllabic, and lexical representations emerged in both CNN- and RNN-based models trained on the same data; (ii) phonemic information was most prominent in the early layers, syllabic information in the middle, and lexical information in the deeper layers; (iii) the results were qualitatively similar for synthetic and real speech; and (iv) the findings generalized to real infant-directed speech from the Brent corpus (Brent & Siskind, 2001), which was not used in model training. In addition, Khorrami and Räsänen found (v) substantially above-chance temporal segmentation ability for the three representation types emerged in the network. Later, Peng and Harwath (2022) and Peng et al. (2023) also found lexical and syllabic segmentation as an emergent property of cross-situational learning in VGS models. Together, these results support the hypothesis of Räsänen and Rasilo (2015) that a separate word segmentation mechanism may not be necessary to bootstrap protolexical learning, but segmentation can emerge from holistic meaning-centric processing of longer auditory patterns (cf. Tomasello, 2000).

Recently, Khorrami et al. (2023) investigated a combination of SSL-based auditory learning and VGS-based audiovisual learning using a model architecture adapted from Peng and Harwath (2022). They explored the developmental timeline of phonemic, lexical, and word-referent learning under different balances of auditory and audiovisual training—for example, beginning with auditory learning followed by audiovisual learning, maintaining a strong audiovisual emphasis throughout, or training with an equal emphasis on both training mechanisms. As expected, auditory-only learning failed to capture word-referent knowledge but otherwise matched the findings of Lavechin et al. (2025), reaching high phonemic discrimination performance and clearly above-chance lexical discrimination. When audiovisual learning followed auditory learning, the rate of lexical learning substantially increased, and word-referent knowledge began to emerge rapidly.

Khorrami et al. (2023) also discovered two unexpected findings: First, when combined audiovisual and auditory learning was followed by auditory-only learning, the model remained highly robust against catastrophic forgetting, including for referential knowledge.

Since humans also receive much of their speech input without immediate visual referents, maintaining robust representations under varying exposure to uni- and multimodal learning contexts is likewise desirable for computational models. Second, the relative order of acquisition for phonemic, lexical, and referential knowledge remained consistent, regardless of whether learning was driven by auditory or audiovisual predictive processing. In other words, even when the learner prioritized cross-modal grounding of speech patterns, phonemic knowledge always emerged first, followed by lexical discrimination and, finally, word–referent pairings. This demonstrates that observable developmental trajectories in language acquisition do not require the learner to focus on any particular aspect of speech, where superficially similar outcomes can arise from different learning strategies and statistical processes.

As for other studies with VGS models, Kamper et al. (2019) used a model architecture that uses a speech encoder to explicitly predict visual objects recognized by the visual encoder. By first training the system with 30,000 pairs of utterances and images, they then investigated how audiovisual semantics emerge in the model. They compared the model's semantic relatedness ratings of image-utterance pairs to human ratings collected for the same pairs, finding that the pairs humans found semantically related were also systematically considered as related by the model. Merkx et al. (2019) improved the audiovisual speech-to-image retrieval performance of the earlier VGS models and also showed that word knowledge is most prominent in the deeper layers of the speech encoder network, replicating earlier findings. Nikolaus et al. (2022) studied VGS learning using Peppa Pig cartoons as model input. Their study demonstrates how a VGS model can acquire word and utterance meanings from the cartoon contents despite the notable referential ambiguity and temporal asynchrony between contents of the speech and video streams.

**Time-course and competition in word activation of VGS models**

A few studies have investigated the time course of activations inside VGS models during speech perception. Havard et al. (2019b) used a gating paradigm (see Grosjean, 1980) to examine how words activate and compete with each other in a trained VGS model. By feeding an RNN-based VGS model with isolated word tokens whose duration increased gradually from the beginning (or end) in small steps—that is, with progressively longer partial words—they observed that word onsets were much more important for successful word recognition than word endings. This finding aligns with the special status of word onsets in human word recognition (e.g., Marslen-Wilson & Zwitserlood, 1989). However, analysis of word competition during ambiguous onsets (e.g., plate vs. player) showed that, unlike human listeners, the activation patterns related to retrieving a specific visual referent did not reflect multiple equally strong concurrent hypotheses in the active cohort. Instead, the network tended to commit to a single hypothesis at each moment in time, rapidly switching to an alternative hypothesis when conflicting evidence unfolded over time.

Another study investigating the temporal dynamics of word activation and competition in VGS models was conducted by Merkx et al. (2023). Using real rather than synthetic speech in a gating paradigm, they used linear mixed models to investigate acoustic and linguistic factors related to word recognition accuracy of the model. They found that noun recognition was more difficult for a larger word-initial cohort of competing words but became more accurate as increasingly many phones were observed. They also observed a negative interaction between the neighborhood density of the input words (the number of phonologically similar words) and their frequency during learning—even though density alone was a positive predictor. The authors interpreted this interaction as reflecting two competing effects: high neighborhood density (many similar words) during learning supports recognition of low-frequency words by providing more perceptual learning opportunities across the occurrences of similar sounding words, whereas high density becomes detrimental when there would otherwise enough exposure to learn a word, but where the lexical competition adds inherent ambiguity to the recognition process. This interpretation aligns with empirical evidence from human studies (Metsala, 1997; Goh et al., 2009). Merkx et al. (2023) also found that the VGS model had learned to distinguish singular from plural nouns and generalized to isolated words despite being trained on continuous utterances. However, the model did not learn verbs effectively, likely because extracting actions from static images is much more difficult than from dynamic visual data such as video.

### Summary of VGS models

Overall, studies on VGS models indicate that visual grounding of spoken utterances is a robust mechanism for acquiring linguistic knowledge across multiple representational levels. It provides access to speech segmentation without the need for any explicit segmentation mechanism (cf. Räsänen & Rasilo, 2015), and the learning outcomes appear largely independent of the specifics of the input data or the architectural details of the deep neural network. Moreover, once trained, VGS models exhibit several similarities to human word recognition processes. It is also worth noting that the general structure of neural VGS models resembles that of the RNN-based EARSHOT model of adult speech recognition (Magnuson et al., 2020). However, whereas EARSHOT requires ecologically implausible supervised training with lexical targets, VGS models learn to recognize words and their meanings by tracking cross-situational statistics in audiovisual input without any linguistic labels or other strong priors.

VGS models also display learning trajectories in which phonetic knowledge tends to emerge first, followed by the development of word discrimination ability and word–meaning associations. This occurs even when the learning objective is visually grounded word learning from the outset. On the surface, this pattern aligns with the classical view that language acquisition begins at the segmental level through perceptual attunement to the native-language sound system, subsequently enabling lexical learning (cf. Kuhl, 2004). However, the VGS models suggest that the skills emerging first need not be explicit learning targets. Instead, they may represent emergent “prerequisite” outcomes that support the efficient acquisition of meaning-centric predictive processing.

## Recent advances in improving the realism of learning simulations

### Improving environment models for learning simulations

One major limitation in most existing computational studies on statistical learning is the limited ecological plausibility of the speech input data. Many of the reviewed studies have used either crowd-sourced audiobooks of read speech or spoken or synthesized image captions, both consisting of speech targeted at adult listeners. In contrast, real infants are primarily exposed to child-directed speech (CDS) and adult-to-adult conversational speech across a variety of everyday contexts, with CDS often tailored to the recipient’s language proficiency. Real speech input also predominantly comes from primary caregivers, siblings, or other close members of the social environment, details depending on the family and sociocultural context. In most simulations, however, the proportions of speech from different speakers are not matched to real-world conditions. Finally, infants' language experiences are characterized by highly variable auditory environments, where a large proportion of input is mixed with substantial environmental sounds, resulting in a typical average speech-to-noise ratio (SNR) in the range of 0–10 dB (see, e.g., Lavechin et al., 2024; Räsänen et al., 2019; see also Coffey et al., 2024)—a condition not reflected in audiobook or image-caption corpora.

***Learning from child-centered audio recordings.*** To address the limitations in the representativeness of speech input, Lavechin et al. (2024) investigated how SSL-based auditory phonetic learning performs when the learner is exposed to real-world speech heard by infants. They trained learner models using the same STELA SSL architecture as in Lavechin et al. (2025; see Section 3) on English and French long-form recordings collected from children's home environments with wearable recorders. Phonetic discrimination was then measured using the ABX task in both languages. The results showed that learning from real-world child-centered data was considerably more challenging than learning from audiobooks. Specifically, the model did not exhibit the same native-language advantage in phonetic discrimination as a model trained on audiobook data.

However, when Lavechin et al. introduced several inductive biases into the learner model, some, but not all, of the native-language advantage was regained. These biases included the ability to separate speech from non-speech input, distinguish between speakers, and the use of data-augmentation techniques to increase variability in the linguistic exposure. Overall, their work represents an important step toward more realistic training data for computational simulations, showing that suitable inductive biases can support learning from complex real-world input. At the same time, it also demonstrates that current SSL models may not yet be fully equipped to handle the challenging real-world auditory scenes, in which the human brain engages in extensive non-linguistic auditory processing (e.g., binaural hearing, auditory scene analysis, selective attention, dynamic cochlear filtering; see, e.g., Bregman, 1990). Currently, such mechanisms remain largely absent from existing computational models.

***Creating controlled but realistic speech input.*** The work by Lavechin et al. (2025) highlights the difficulty of dealing with noisy, real-world long-form audio. It is also challenging to disentangle how different aspects of the data contribute to the model's learning outcomes. For instance, is the poorer performance compared to audiobook simulations due to the low SNR, or to the distinctive characteristics of caregiver speech in natural interactions? How does the distribution of speakers influence learning? Is CDS easier or harder to learn from than other speech types? And how do these factors interact? Moreover, even the densest long-form data from individual children fail to capture the entire language experience of these children, making it difficult to systematically examine how variations in input quality and quantity across families affect learning outcomes.

To address these challenges, Räsänen and Kocharov (2025) proposed an environment modeling approach in which the learner's speech input is generated synthetically. Using an LM and text-to-speech synthesizer, their approach allows production of large quantities of speech statistically similar to real caregiver speech, enabling controlled and systematic investigations of how different experiential factors influence learning. Their approach allows comparisons between simulations that differ in only a single dimension—for example, keeping the linguistic content constant while changing the speaking style from neutral to more expressive prosody, or maintaining the same speaking style while switching the content type from CDS to storybook narratives. It also allows tailoring of the input to the age of the addressee child, enabling simulations with appropriate (or intentionally varied) input quality and quantity at different points along the developmental timeline.

***Modeling prenatal language experience.*** Another aspect of input relevant to ecological plausibility is the effect of prenatal language exposure on early language development. The auditory system of the human fetus becomes operational during the last trimester of pregnancy (e.g., Holst et al., 2005), and Gervain (2018) has suggested that fetuses not only have access to speech in the womb but that their brains may begin to adapt to the prosodic properties of that input before birth. More broadly, although speech is heavily attenuated in utero and the auditory system is still developing, the possibility of learning about speech before birth could help explain certain early language abilities observed in newborns, such as preference for the mother's voice, vowel discrimination, and language discrimination across rhythmic families.

To investigate this prenatal bootstrapping hypothesis computationally, Cruz Blandón et al. (2025) used APC and CPC-based statistical learner models to test how varying amounts of prenatal speech exposure might affect later language-learning outcomes. Drawing on empirical data on fetal hearing and sound attenuation in the womb, they implemented an acoustic filter simulating the auditory experience of fetuses during the last trimester of pregnancy. The APC and CPC models were then trained to perform statistical learning up to 12 months of corrected postnatal age, using filtered (prenatal) and full-band (postnatal) speech input. Matching the total exposure to empirical estimates of infants' real-world language input, they compared simulated full-term and preterm infants to a baseline model without prenatal exposure.

Cruz Blandón et al. evaluated the learner models on infant-directed speech preference, vowel discrimination, phonotactic preference, and tone discrimination. The latter three abilities had been empirically tested by Gonzalez-Gomez et al. (2021) in preterm and full-term infants, showing that tone discrimination was developmentally delayed in preterm but not full-term infants. The question was whether including prenatal exposure in the computational simulations would replicate this behavioral pattern. The results showed that prenatal exposure did not generally improve overall model compatibility with empirical infant data across all capabilities and ages. However, it substantially altered the learning trajectories of these skills as a function of the amount of prenatal exposure. This suggests that prenatal experience can modulate learning and should be incorporated into comprehensive computational models of early language development.

***Towards realistic VGS model training.*** In the context of VGS models, a major issue in prior studies has been the massive scale of audiovisual data used for training. For example, the SPEECH COCO dataset (Chrupała et al., 2017) and the Places Audio Captions (English) 400k dataset (Harwath et al., 2016) each contain hundreds of thousands of images paired with spoken descriptions, corresponding to hundreds of hours of speech that are highly descriptive of the concurrent visual input. While real infants indeed have continuous access to parallel auditory and visual information, the proportion of caregiver speech that explicitly refers to concrete objects and events within the infant's visual field is likely to be substantially lower than in these experimental datasets (cf. Clerkin & Smith, 2022).

To examine whether VGS models can learn from a realistic amount of audiovisual experience relevant to cross-situational word learning, Khorrami and Räsänen (2025) conducted a study that matched the statistical properties of VGS training data with empirical estimates of audiovisual word–referent co-occurrences. Drawing on visual object naming rates reported by Clerkin and Smith (2019, 2022), they created a modified version of the SPEECH COCO dataset simulating daily naming frequencies for the 80 most common objects, aligned with observed empirical rates. The total number of naming events was then extrapolated to represent 2, 4, or 6 months of audiovisual learning. Because both the objects and words appeared within larger visual scenes (photographs) and spoken captions, substantial referential ambiguity remained in the data—consistent with naturalistic conditions. Instead of trying to learn from audiovisual experiences directly, a preceding auditory-only learning phase was included to simulate the first six months of language development, whereas the employed self-supervised visual encoder was taken as an approximation of visual perceptual learning during the same period. This was done to approximate perceptual learning in the individual perceptual modalities before the learner obtains sufficient motor skills for upright posture (e.g., sitting) and active exploration of the world, thereby enabling more systematic and stable audiovisual experiences in everyday communicative situations. Using a learner model that combined SSL and VGS learning, they investigated phonemic discrimination, acoustic word-form discrimination, and audiovisual word–referent mappings in a simulated infant at 6, 8, 10, and 12 months of age.

As a result, Khorrami and Räsänen found that the model successfully bootstrapped lexical learning from the limited but ecologically realistic number of audiovisual naming events, both in terms of word discrimination and word–referent mapping. Furthermore, the growth of the receptive lexicon was qualitatively similar to that of real infants across the studied age range. Notably, lexical learning was only successful when audiovisual training followed the initial auditory learning phase, but not when audiovisual learning was attempted directly from the limited number of naming events. This finding suggests that early auditory perceptual learning may be essential when multimodal learning opportunities are sparse, as it stabilizes the auditory representations needed to support the systematic tracking of cross-situational statistics.

As for other studies attempting to improve the realism of environment modeling, Räsänen and Khorrami (2019) applied an audiovisual learner model to acoustic speech and visual referent data obtained from head-mounted cameras on infants, obtained from the SEEDLingS corpus (Bergelson et al., 2019). After training the model with approximately 3.5 hours of visually grounded speech, the learner began to show signs of emerging lexical and phonetic knowledge. However, their setup was simplified in that it did not use the actual images from infants' visual fields but relied instead on manually annotated labels describing what the infant was attending to in each scene.

As a complementary approach, Vong et al. (2024) recently used longitudinal head-mounted camera data from a single infant, this time including real images of the infant's visual field but replacing the acoustic input with transcripts of the recorded speech. Their results showed that if the learner has access to an invariant discrete representation of spoken language[3], the acquisition of numerous word–referent mappings is possible from infants' own visual experiences.

To date, however, modeling studies that simultaneously use real child-centered audio and visual input at a realistic scale are still missing, though such work will likely become feasible as denser longitudinal and higher-quality audiovisual datasets from infants' everyday lives together with increasing computing power become available.

**Improving ecological relevance of outcome models**

As the final topic of this chapter, we turn to recent advances in evaluating computational models of early language development. As discussed earlier, each modeling study requires three essential components: the environment model, the learner model, and the outcome model. The outcome model serves to determine what the learner has acquired and how those learning outcomes relate to language development in real infants. When an outcome model establishes a clear connection between computational results and empirical developmental phenomena, it can be considered ecologically relevant.

The classical approach to evaluating computational models is to compare their learned representations or behaviors against a linguistic ground truth. This ground truth typically relies on manual annotations of speech data, such as phonetic segment labels for testing phoneme discrimination (cf. the ABX test; Schatz et al., 2013) or phone/syllable/word boundaries for assessing speech segmentation (e.g., Räsänen et al., 2018; Scharenborg et al., 2007). By feeding the model with speech input and comparing the resulting internal representations or acceptability judgments to the annotated reference, researchers can obtain performance scores for different levels of linguistic representation.

Comparison of models against linguistic annotations is highly informative for determining what the model has or has not learned, and it therefore serves as the foundation for producing learnability proofs in modeling studies. However, studies using this approach often implicitly assume that the best model is the one achieving the highest possible score in the given benchmark. In other words, the model that parses speech most similarly to how a trained linguist would analyze it is taken to be the most successful learner. While this may serve as a reasonable proxy for adult linguistic competence (at least for literate adults; see Port & Leary, 2005), infants are not small adults. Their language skills undergo continuous development, and it remains unclear what types of internal representations support the observable linguistic behaviors of children at different developmental stages. Indeed, one of the key contributions of computational modeling is precisely to explore what kinds of intermediate representations could give rise to those behaviors.

Another limitation of many evaluation approaches parallels the plausibility concerns raised for environment models. If the annotated speech data used to probe a model's competence differs substantially from the speech input encountered by real learners in their everyday environments, the resulting benchmarks may fail to reflect the model's actual functional capabilities in relevant communicative contexts.

[3] But see the introduction of this chapter for limitations in using discrete input representations for models of early learning.

***Modeling lexical and syntactic learning.*** To better align model evaluation with infant language development in natural settings, several important advances have recently emerged. Drawing inspiration from the BabyLM benchmark for language models (Warstadt et al., 2023), Lavechin et al. (2023) introduced a so-called BabySLM benchmark to assess the lexical and syntactic skills of speech- and transcript-based models of early development. For lexical evaluation, the model is asked to provide acceptability scores for isolated words that occur in CHILDES transcripts (MacWhinney & Snow, 1985), thereby approximating the vocabulary actually heard by real infants. These scores are then compared against those for corresponding minimal-pair pseudowords to determine whether the model assigns higher acceptability to true lexical items (see Section 3.1). The resulting lexical knowledge score reflects the model's specificity in encoding native-language vocabulary. For syntactic testing, the model is presented with utterances containing particular syntactic constructions alongside their agrammatical minimal pairs, such as "*The good mom*" versus "*The mom good*" for adjective–noun order. The model is expected to yield higher acceptability ratings for the grammatical alternatives (see also Warstadt et al., 2020, for a similar method applied to text-based systems).

Another test for early acoustic word-form discrimination was introduced by Khorrami et al. (2023). In this CDI-Lextest, the model is presented with isolated words drawn from the 89 lexemes listed in the English CDI short forms (parental questionnaires used to assess infants' early vocabulary). The words are synthesized in multiple voices and prosodic variations and fed to the model as isolated tokens. The resulting model representations are then analyzed for how consistently different tokens of the same word type cluster together in the representation space, and an overall word-form discrimination score is derived. In contrast to BabySLM, which evaluates broader and more fine-grained lexical encoding across a wider developmental span, CDI-Lextest focuses on lexical discrimination abilities that are immediately relevant to early communicative contexts, such as distinguishing "*book*" from "*bird*" or "*dog*" from "*doll*".

***Modeling developmental trajectories.*** Since infant language development is in constant flux, an accurate computational model should also reproduce the developmental trajectories observed in real infants. Beyond CDI-style lexical questionnaires, trajectories for various language capabilities are primarily derived from empirical in-lab studies conducted with infants of different ages. However, such studies typically cover only a limited range of age groups, owing to practical and logistical constraints. At the same time, the replicability of individual findings has become a persistent concern in developmental research for more than a decade (e.g., Open Science Collaboration, 2015). In response, the field has increasingly turned to large-scale collaborative efforts (e.g., ManyBabies Consortium, 2020) and meta-analytic approaches to development (e.g., Bergmann et al., 2018) to integrate and interpret data from multiple studies, and instead of relying on individual findings. Meta-analyses are particularly valuable for deriving developmental trajectories, as infant age can be treated as a moderator variable explaining the variance observed across studies targeting the linguistic capability of interest.

To benchmark the developmental trajectories of computational models against those of real infants, Cruz Blandón et al. (2023) introduced a meta-analytic evaluation framework called MetaEval. This approach uses a meta-analytic model of empirical data on a given infant language capability as the reference and then derives a corresponding developmental trajectory for the computational model, treating the amount of speech input as a proxy for simulated age. To obtain the model's developmental trajectory, the model is exposed to computational replications or approximations of the original empirical paradigms used to test infants—for example, native versus non-native vowel discrimination tasks employing the same phonetic contrasts as in behavioral experiments. As in empirical meta-analyses, a statistical model is then fitted to test outcomes to estimate model competence as a function of simulated age. Comparing the resulting model and human developmental trajectories then reveals whether the model reproduces infant-like behavior across the simulated age range.

To advance the development of holistic models of early language learning, MetaEval also promotes parallel measurement of multiple language capabilities using the same environment and learner models. In this view, a robust model of early language development should replicate infant-like developmental trajectories across a broad range of linguistic phenomena. Testing multiple capabilities in parallel further mitigates the risk of overfitting (over-designing) a model to explain one isolated aspect of language while failing to generalize to others. It also removes the issue of a model being incompatible with other models of learning addressing different aspects of language. Importantly, this multi-capability evaluation does not depend on whether the model being tested comprises a single or multiple learning mechanisms—the only requirement is that these mechanisms operate in concert on shared input to produce the learning outcomes observed in infants, without modifying model parameters between tasks.

Recently, Tan et al. (2024) proposed DevBench, a framework for systematic benchmarking of audiovisual computational models of language learning against behavioral data from children. Like MetaEval, DevBench compares human experimental data with model performance on corresponding tasks, encompassing assessments of lexical, syntactic, and semantic skills. However, in contrast to MetaEval that relies on effect sizes aggregated from multiple empirical studies, DevBench is based on item-level human response data obtained from diverse experimental paradigms, such as looking-while-listening and visual vocabulary tasks assessing lexical knowledge. By comparing the distributions of response scores between human participants and a model across identical test items, the framework yields a human-data compatibility score. Calculating this score across simulated and real age groups further allows the derivation of developmental trajectories for the tested language capabilities.

Since the experimental paradigms of DevBench make use of audiovisual processing, the models that can be evaluated with it must also be capable of (relative) scoring of visual inputs in the context of speech input, making the benchmark suitable for evaluating VGS models of learning. At the time of its publication, the age range of behavioral reference data in DevBench primarily consisted of older children instead of infants. Nonetheless, the general framework is readily extensible to early language development, provided that suitable infant behavioral data—and possibly complementary infant-appropriate experimental protocols—are incorporated into the benchmark.

## Summary

This chapter has provided an overview of modern computational approaches to studying early language development in infants, including recent advances in modeling the learning environment and evaluating learning outcomes with respect to infants' linguistic development. We acknowledge that the review is not exhaustive: many relevant and influential studies—particularly earlier works and those employing computational approaches other than SSL or VGS—were not included. The primary goal has been to offer a concise synthesis of recent progress in speech-based and audiovisual models of early language learning. For example, models addressing speech production (e.g., Georges et al., 2024; Guenther, 1995; Warlaumont & Finnegan, 2016), language learning in embodied robots (see Oudeyer et al., 2019, for review), and approaches operating on non-acoustic representations of spoken language were not addressed despite their central relevance for a complete understanding of early language acquisition.

In terms of theoretical implications, the reviewed computational models of auditory and audiovisual learning demonstrate that many aspects of early language acquisition can emerge without strong linguistic priors or explicit inductive biases. Rather than aiming to learn specific linguistic units or structures, these modern models acquire language-related capabilities as a by-product of generic predictive optimization operating within and across sensory modalities. By doing so, they provide compelling evidence that the extraction of statistical regularities from sensory input constitutes a fundamental mechanism in early language development.

At the same time, these models extend beyond classical statistical learning frameworks, such as those focused on tracking syllable-transition probabilities for word segmentation (e.g., Saffran et al., 1996) or learning distributions of acoustic cues for phonetic category formation (Maye et al., 2002). They do so by bypassing the need to predefine elementary units or specify the exact statistics to be learned. Instead, they autonomously discover what needs to be represented in order to minimize surprisal in the unfolding sensory input—enabled by the flexibility of deep neural networks to learn complex nonlinear predictive mappings. In other words, these models develop hierarchically organized latent representations of varying granularity and interdependence in a fluid, incremental manner, as long as these representations support the overarching predictive task within the sensorimotor environment. This property makes them not only compatible with predictive processing theories of the human brain but also parsimonious, requiring relatively few explicit learning mechanisms to account for the emergence of early linguistic knowledge.

The discussed models are also broadly compatible with several theoretical frameworks of early language development. They begin with general perceptual and holistic processing of utterances, from which phonemic and lexical knowledge gradually emerge through experience with speech and through exposure to how speech is used to communicate about the world. This progression aligns with the principles of PRIMIR (Curtin et al., 2011; Werker & Curtin, 2005) and with usage-based theories of language development (Tomasello, 2000). At the same time, even though these models are not explicitly designed to learn phonemic representations, they exhibit signs of perceptual reorganization for phonemic perception before lexical knowledge becomes prominent, thereby reproducing the classical developmental timeline observed in infants (cf. Dupoux, 2018, Kuhl, 2004, and references therein). Moreover, the incremental and predictive nature of learning in these systems are broadly compatible with several of the key principles of constructivist theories (e.g., Rowland et al., 2026): the ability to learn from multimodal input, existing internal representations impacting processing of new input, continuously adapting to the external environment through predictive, experience-driven updating of internal models.

Despite these advances, it is important to acknowledge that current computational models capable of learning from real speech remain substantially limited as models of infant learners. The existing simulation setups represent only simplified reflections of real-world learning and infant–caregiver communication. All the reviewed studies focus on perceptual processing of input unfolding in a predetermined manner, independent of the learner's actions. This neglects the interactive and exploratory dimensions of early development, including how caregivers adapt their speech and communicative behavior to the child's situational context and developmental level.

Furthermore, the models fall short as cognitive models, since they are relatively direct adaptations of their original machine learning architectures, lacking the biological and cognitive constraints that shape real human learning. They also typically achieve their best performance when trained iteratively across multiple passes over the data—a property of current artificial neural networks but one that is implausible for human learners (see also Cruz Blandón et al., 2025). In addition, there have been few attempts to model individual variability in development, as most models are interpreted as average learners, both in terms of learner properties and in relation to the aggregated empirical data used for comparison. Finally, substantial work remains to be done in developing more realistic yet controllable environment models and in systematically benchmarking model outcomes against empirical infant data, including efforts to incorporate multilingual learning into simulations.

Finally, the connection between current speech learning models and formal theories of development and cognition also remains somewhat superficial. While the current models generally adhere to several theoretical principles at a high level, there is also notable variability in how predictive models of speech learning can be implemented, and none of the reviewed models have been designed as direct computational implementations of specific scientific theories of early language development. While this flexibility demonstrates that high-level theoretical frameworks are often too general to constrain specific computational mechanisms, it also highlights the need for a closer integration between computational modeling and empirical developmental research. A tighter exchange of hypotheses, ideas, and findings between the two domains would likely accelerate progress toward mechanistically grounded, developmentally realistic, and holistic models of early language acquisition.